\documentclass[final]{cvpr}

\usepackage{times}
\usepackage{epsfig}
\usepackage{graphicx}
\usepackage{amsmath}
\usepackage{amssymb}
\usepackage{booktabs}
\usepackage{caption}

\usepackage[pagebackref=true,breaklinks=true,colorlinks,bookmarks=false]{hyperref}

\def\cvprPaperID{****} 
\def\confYear{CVPR 2021}

\begin{document}

\title{Understanding and Improving Group Normalization}

\author{Agus Gunawan, Xu Yin, Kang Zhang\\
{\tt\small \{agusgun, yinofsgvr, zhangkang\}@kaist.ac.kr}
}

\maketitle

\begin{abstract}
Various normalization layers have been proposed to help the training of neural networks. Group Normalization (GN) is one of the effective and attractive studies that achieved significant performances in the visual recognition task. Despite the great success achieved, GN still has several issues that may negatively impact neural network training. In this paper, we introduce an analysis framework and discuss the working principles of GN in affecting the training process of the neural network. From experimental results, we conclude the real cause of GN's inferior performance against Batch normalization (BN): 1) \textbf{unstable training performance}, 2) \textbf{more sensitive} to distortion, whether it comes from external noise or perturbations introduced by the regularization. In addition, we found that GN can only help the neural network training in some specific period, unlike BN, which helps the network throughout the training. To solve these issues, we propose a new normalization layer built on top of GN, by incorporating the advantages of BN. Experimental results on the image classification task demonstrated that the proposed normalization layer outperforms the official GN to improve recognition accuracy regardless of the batch sizes and stabilize the network training.
\end{abstract}

\section{Introduction}
Throughout the development of deep learning, Batch Normalization (BN) \cite{ioffe2015batch} plays a vital role. It normalizes the feature maps using the mean and variance computed within a (mini-)batch. Various studies have shown BN's remarkable effectiveness in easing neural network training and help deeper networks to converge \cite{he2016deep, huang2017densely, szegedy2016rethinking}. Despite the great success it achieved, BN fails in some cases when the batch size is small \cite{ioffe2017renormalization, wu2018group, lian2019revisit}. This is because a small batch size can give an inaccurate estimation of statistics information that worsens the normalization layer's performance. To tackle this problem, there have been many studies proposed, such as Instance Normalization (IN) \cite{ulyanov2016instance} and Layer Normalization (LN) \cite{ba2016layer}. These layers can work under small batch size setting and proved to be effective for training generative model \cite{goodfellow2014generative} and sequential model \cite{hochreiter1997long} correspondingly. However, we still need an alternative to BN in visual recognition tasks. Among amounts of variants, Group Normalization (GN) \cite{wu2018group} is well-known in the community and shows a better and stable performance in the recognition task. However, it is also reported \cite{wu2018group} that GN has inferior performance compared to BN in large batch size settings. Although the authors \cite{wu2018group} gave a simple explanation where this problem results from the lack of regularization in the GN layer, we argue that this issue is not the root cause of this problem.

In this study, we show some real factors of GN's inferior performance against BN and make a detailed explanation of GN's working principles. To help the analysis, we employ two numerical metrics, loss landscape and gradient predictiveness, which are used to "describe" and evaluate the normalization effects in the training process (a similar work can be referred to \cite{santurkar2018how}). Apart from the analysis, we also design a new normalization layer built on top of GN by incorporating the mechanism of BN to GN. To the best of our knowledge, our work is the first attempt to analyze the issues of GN and further make improvements on GN. The code for the analysis experiments and proposed method can be seen in our repository\footnote{\href{https://github.com/agusgun/UnderstandingGroupNorm}{https://github.com/agusgun/UnderstandingGroupNorm}}.

In summary, our contributions are listed as follows:
\begin{itemize}
\item We analyze the Group Normalization layer's behavior in the network training by using an analysis framework. In addition, we examine GN's robustness and compared it with BN in different cases, e.g adding external noise and introducing perturbations.
\item Inspired by other studies, we design a new normalization layer built on top of GN by integrating BN's mechanism into the GN layer to solve the aforementioned issues and therefore have better performance compared to a vanilla GN.
\item Finally, experiment results on visual recognition tasks (evaluated on \textbf{CIFAR-10, CIFAR100, and SVHN}) demonstrate that our new normalization layer is better than the baseline method (the original GN) in improving networks' performance and stabilize the training, without influences from the batch sizes.
\end{itemize}

\section{Related Works}
\noindent\textbf{Normalization layer.} Batch Normalization (BN) is one of the most fundamental components in the development of deep learning. By computing the mean and variance within a (mini-)batch, BN enforces normalization on the feature maps, which can ease the training of the deep neural network and help the network to converge \cite{ioffe2015batch}. However, various analyses and studies pointed out that BN is suffering from its inaccurate statistic estimations when the batch size becomes small \cite{ioffe2017renormalization, wu2018group, lian2019revisit}, indicating that it needs a sufficiently large batch size to work.

To address this weakness of BN, many studies have been proposed to improve this layer which are Layer Normalization (LN) \cite{ba2016layer} and Instance Normalization (IN) \cite{ulyanov2016instance}. LN  calculates the statistic (mean and standard deviation) over all hidden units in the same layer. Meanwhile, IN calculates the statistics for each sample and each channel. The latest proposed normalization layer is Group Normalization (GN) \cite{wu2018group}, which is effective in the visual recognition task compared to IN and LN. In this study, we choose to analyze various issues in GN and put an effort to improve this layer by incorporating a mechanism from BN that can improve GN's effectivity in neural network training. \\

\noindent\textbf{Understanding how the normalization layer helps optimization.} 
In the early time when Batch Normalization (BN) is first proposed, the normalization layer is believed to help optimization because it reduces the internal covariate shift \cite{ioffe2015batch}. However, the study in \cite{santurkar2018how} shows that there is no relation between internal covariate shift and the effectiveness of BN in helping the optimization of the neural network. The study shows that the effectiveness of BN is because it makes the optimization (loss) landscape more smooth. As the effect, it makes the training becomes more stable and faster. The work uses various empirical studies using loss landscape \cite{li2018visualizing} and gradient predictiveness to analyze the impact of the normalization layer on the stability of the optimization and network training. In this work, we use the same metrics used in \cite{santurkar2018how}, consisting of loss landscape and gradient predictiveness to analyze the issues in GN related to network training. \\

\noindent\textbf{Existing improvements on GN.}
In recent years, various studies have been proposed to improve the normalization layer \cite{ioffe2017renormalization, summers2020four, zhou2020batch, jingjing2019understanding, huang2017arbitrary}. In \cite{ioffe2015batch}, two main issues that BN has are that small mini-batches can provide inaccurate statistics estimation, and non-i.i.d. mini-batches have a detrimental effect on the network with BN. To solve these issues, the work proposes batch renormalization. The study in \cite{summers2020four} also tries to improve BN by incorporating four improvements. One of them is General Batch and Group Normalization (GBGN). It uses a grouping mechanism in the batch dimension. Meanwhile, the study in \cite{zhou2020batch} proposes Batch Group Normalization (BGN) that also tries to incorporate a grouping mechanism, but in the channel, height, and width dimension. Other works try to improve other normalization layers, such as Adaptive Instance Normalization (AdaIn) \cite{huang2017arbitrary} that improves IN for style transfer, and AdaNorm \cite{jingjing2019understanding} that improves LN by controlling scaling weight adaptively towards different input. 

To the best of our knowledge, there is no existing work that focuses on analyzing the issues in GN and improving this layer. Our idea of incorporating BN calculation to GN has some similarity to GBGN \cite{summers2020four} and BGN \cite{zhou2020batch}. However, their work incorporates a grouping mechanism to improve BN. Meanwhile, our method incorporates BN calculation adaptively to improve GN.

\section{Issues in Group Normalization (GN)}
The original study of GN \cite{wu2018group} explains that the main reason for GN's inferior performance against Batch Normalization (BN) is the lack of regularization that BN has naturally in its mechanism \cite{luo2018towards}. However, we find that both distortions of the output distribution and noise introduced by regularization can worsen neural network training that uses GN. In addition, we also find that GN's effectiveness in helping neural network training is worse compared to BN. 

\subsection{Analysis Framework}
To start with, we employ a similar analysis framework in \cite{santurkar2018how}, which is used to analyze BN's behavior in the training process. First, we analyze the \textbf{loss landscape} \cite{li2018visualizing}, which shows the landscape of the optimization (loss value) through the training process. Specifically, for each step in the training process, we measure how the loss changes at each step after moving to different points along the direction of the calculated gradient. The second metric introduced here is \textbf{gradient predictiveness} which aims to measure the stability and predictiveness of the gradients. It is calculated by the $\ell_{2}$-distance between the loss gradient at the given point (step) and the previous point (step) in the training process. A lower value in the loss landscape and gradient predictiveness, and a smaller range of values for both these metrics denote better training performances.

The network used throughout the analysis experiment is VGG-11, which can be trained without any normalization layer \cite{simonyan2015vgg}. In addition, we also found that GN can be used across different network families (shown in Appendix \ref{app:exp_gn_aplicability}), which means the choice of the network will not affect the analysis. To help the analysis, we also compare the loss landscape and gradient predictiveness of GN and BN in the experiments. We choose BN as the baseline because this layer is known to help the training process shown in \cite{santurkar2018how}. The detail of the analysis experiment can be seen in Appendix \ref{app:analysis_experiments_details}.

\subsection{Analysis Result}
\label{sec:analysis_result_issues_in_gn}

\subsubsection{GN Sensitivity to The Distortion of Output Distribution}
Before analyzing GN's impact on the network training, we have to understand the effect of the noise that can distort or even change the output distribution of the GN layer. This condition may cause the internal covariate shift to occur \cite{santurkar2018how}. In this part, we analyze the aforementioned distortion on the training performance of the network with GN. To facilitate this, we add Gaussian noise with $\mu = 10^{-3}$ and $\sigma = 1.001$ for the output of the normalization layer. As shown in Figure \ref{fig:appendix_gn_with_distortion}, the distortion in the output distribution of the GN layer can disrupt the optimization of the network with GN greatly. As a result, the network suffers from a gradient vanishing problem, which makes further training futile. In contrast, the network that uses the BN layer avoided this problem, and even achieves the best validation performance compared to the network with GN. It demonstrates that \textbf{GN is very sensitive to distortion on the output distribution of this layer}. Based on this observation, we conclude that the grouping mechanism of the GN layer must be done correctly, or the underlying mechanism of this layer should be changed to achieve better robustness.

\subsubsection{Impacts of GN on The Network Training}
In this section, we use the loss landscape and gradient predictiveness to measure the impact of GN on the training process of the neural network and report their performance in Figure \ref{fig:appendix_gn_vs_bn_loss_landscape_grad_pred}. We observe that the loss landscape of the network that uses GN has a wider range of values compared to the case of using BN. This wider range shows that the GN's ability to smooth the optimization problem is inferior compared to BN. In addition, the minimum value of the loss landscape when GN is used is constantly higher than the case of using BN. This happens especially in the initial training period, where its minimum value is even higher than the maximum loss value of BN. This fact contradicts the observation in \cite{wu2018group}. The original study shows that GN helps the network to train better compared to BN because of better training performance. However, based on our experiments, GN's impact on network training, especially to smooth the optimization problem, is worse compared to BN's. Furthermore, gradient predictiveness shown in Figure \ref{fig:appendix_gn_vs_bn_loss_landscape_grad_pred} also demonstrates a similar trend as the loss landscape. The wider range of gradient predictiveness of the network that uses GN indicates that GN's effect in easing the network training is less than BN.

To verify this observation, we report the loss landscape of the network with and without GN (vanilla network). The result in Figure \ref{fig:appendix_loss_landscape_grad_pred_gn_vs_nonorm} shows that the loss landscape of the network that uses GN is similar to the vanilla network. In fact, we find that the loss landscape of the vanilla network has a smaller range of values in the early training period compared to the case of using GN. It means that GN actually does not help the network in the initial training period and even can disrupt the training process. Eventually, with the increasing level of optimization, it helps the training (especially in the middle period) to achieve better performance. The worse training performance of GN in the initial period of training actually also has been shown by the plot of the training error in \cite{wu2018group}, where BN only achieves better training performance before the learning rate is decreased using step decay. This fact further supports our observation. We conjecture this observation (GN case) is caused by unreliable feature statistic estimation in the initial training period because the feature rapidly changes in the initial training period (caused by a large update to the network's weight parameters).

In the end, we evaluate the training performance of the network with GN and without GN using gradient predictiveness and show the result in Figure \ref{fig:appendix_loss_landscape_grad_pred_gn_vs_nonorm}. The result shows a similar observation with that of the loss landscape where GN only helps the training of the neural network slightly in the middle period of training. These results indicate that: GN has a higher risk to cause gradient vanishing or gradient explosion compared to BN. It is more sensitive to these issues compared to the vanilla network that does not use any normalization layer in the early period of training. In addition, it also shows that GN's performance depends on the hyperparameter settings, such as learning rate, because of optimization instability shown by the loss landscape.

\subsubsection{Impact of The Regularization Effects on GN}
In this experiment, we follow the suggestion from the study in \cite{wu2018group} to use some regularization terms in the training to increase the validation accuracy of GN. To implement it, we add L2-regularization (weight-decay) in the training process. It is also employed in the GN paper \cite{wu2018group}. However, in our study, we decrease the magnitude of weight decay to $5 \times 10^{-5}$.

\begin{table}[b]
    \centering
    \begin{tabular}{c c c}
    \toprule
    \textbf{Regularization} & \begin{tabular}{@{}c@{}}\textbf{Normalization} \\ \textbf{Layer}\end{tabular} &  \begin{tabular}{@{}c@{}}\textbf{Validation} \\ \textbf{Accuracy}\end{tabular}\\
    \midrule
        - & 
        \begin{tabular}{@{}c@{}}GN \\ BN\end{tabular} & \begin{tabular}{@{}c@{}}\textbf{80.36\%} \\ 82.82\%\end{tabular} \\
    \midrule
        L2-Regularization & 
        \begin{tabular}{@{}c@{}}GN \\ BN\end{tabular} & \begin{tabular}{@{}c@{}}78.34\% \\ \textbf{82.82}\%\end{tabular}\\ 
    \bottomrule
    \end{tabular}
    \caption{The result of using L2-regularization for the network with BN and GN.}
    \label{table:regularization_effect_result}
\end{table}

The result in Table \ref{table:regularization_effect_result} shows that after adding a small regularization, the validation accuracy of the network with GN is significantly reduced (around 2\%). In contrast, the performance of the network with BN is robust to the regularization. It may indicate that the perturbation introduced implicitly by the regularization affects GN's effectiveness in easing network training. This observation is supported by the loss landscape and gradient predictiveness listed in Figure \ref{fig:appendix_regularization_loss_landscape_grad_pred_gn_vs_bn}. By comparing the result of both metrics between the case of adding and without adding regularization, the regularization makes the range of value in loss landscape and gradient predictiveness increased which shows the training performance of GN also becomes worse. It demonstrates that \textbf{GN is sensitive to perturbation induced by the regularization}.

\section{Proposed Method}
The issues of GN in Section \ref{sec:analysis_result_issues_in_gn} mainly show that GN is sensitive to the distortion of output distribution and perturbation introduced by regularization. GN layer also only helps the network training slightly in the middle period of training (e.g. after 1000 training steps). To solve those issues, we have tried to improve the channel-grouping mechanism of GN. However, we find the improved grouping mechanism dramatically increases the training time and complexity, almost ten times of the original version. Therefore, we decide to improve the underlying mechanism of GN by incorporating BN calculation inside the GN layer.

Inspired by \cite{santurkar2018how}, there are two ways to help the training process of neural network (e.g. smoothen loss landscape): incorporating $\ell_{p}$-norm or BN for the normalization. With those inspirations, we decide to incorporate the underlying calculation of BN (i.e. result of BN) to the GN layer since we also want to take the benefit of the regularization effect in BN \cite{luo2018towards}. We also evaluate the proposed idea in \cite{summers2020four} and \cite{zhou2020batch} which uses a similar idea of combining GN and BN. The details of the evaluation of those ideas and naively combining BN and GN (by stacking both of them sequentially) can be seen in Appendix \ref{app:omitted_exp_evaluation_of_other_ideas}.

\begin{figure*}[ht]
\begin{center}
\includegraphics[width=0.95\textwidth]{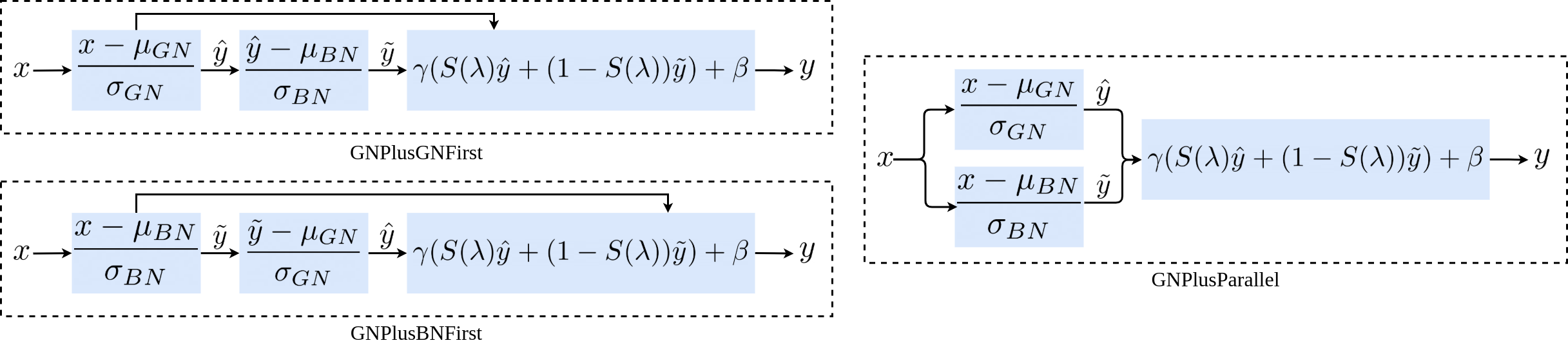}
\end{center}
   \caption{The proposed normalization layer consists of three different variants.}
\label{fig:proposed_method}
\end{figure*}

To start with the explanation, we first formulate the calculation of the normalization layer formally. Suppose $x$ is the output of a layer in the neural network. For 2D inputs, we denote $x$ as a 4D tensor with the dimension of ($N$, $C$, $H$, $W$), where $N$, $C$, $H$, and $W$ denotes batch, channel, spatial height, and spatial width axis respectively. For the normalization, BN computes $\mu_{BN}$ and $\sigma_{BN}$ along the ($N$, $H$, $W$) axes. Meanwhile, GN computes $\mu_{GN}$ and $\sigma_{GN}$ along a group of $C/G$ channels and along the ($H$, $W$) axes. By using this formulation, the proposed method consists of three different new normalization layer variants can be seen in Figure \ref{fig:proposed_method}.

Our main idea is to incorporate the calculation of BN to GN adaptively both in a sequential or parallel manner and the result is controlled by the parameter $\lambda$. Suppose $x$ is the input for all of these new normalization layer variants. The output of GN calculation is denoted as $\hat{y}$, while the output of BN calculation is denoted as $\tilde{y}$. The only difference between these variants is in their inner workings. For sequential variants, we propose two variants which are GNPlusGNFirst and GNPlusBNFirst. Meanwhile, we propose one variant of GNPlusParallel for the parallel variant.

For GNPlusGNFirst, the input is first normalized by using GN calculation. Then, the output of the GN calculation ($\hat{y}$) is normalized by using BN calculation. This means that the output of BN ($\tilde{y}$) consists of two normalization effects induced by GN and BN. Meanwhile, in GNPlusBNFirst, BN takes $x$ as the input and GN uses the output of BN calculation ($\tilde{y}$) as the input. In this case, the output of GN ($\hat{y}$) is the effects of two normalizations of GN and BN. Although in sequential variant the output of later calculation can have two different normalization effects, the later calculation has a more dominant normalization effect. This means the results from these two variants are different because both BN and GN uses ($H$ and $W$)-axes in their calculation, which means the ordering between these two calculations matters. In GNPlusParallel both GN and BN calculation uses $x$ as the input where no output is normalized by two differents normalization sequentially.

After we get the output of GN ($\hat{y}$) and the output of BN ($\tilde{y}$), we will use these outputs in the post normalization process. In this process, $\hat{y}$ and $\tilde{y}$ is multiplied by $S(\lambda)$ and $(1 - S(\lambda))$ where $S(\cdot)$ denotes sigmoid function and $\lambda$ denotes learnable scalar initialized as one. This multiplication represents an adaptive way to incorporate both GN and BN normalization results in the new normalization layer. When the result of GN is reliable for training (when the number of batch size is small), the proposed normalization layer will automatically increase $\lambda$ through training to use GN output only. However, when the result of BN is better to be used for training (when the number of batch size is large), the proposed normalization layer can use the result of BN by decreasing $\lambda$. Similarly as GN \cite{wu2018group} and BN \cite{ioffe2015batch}, we also uses scale $(\gamma)$ and shift $(\beta)$ parameters to enable the transformation to represent identity transform in the proposed normalization layer.

\section{Experiments}
In this experiment, we use ResNet50 \cite{he2016deep} to follow the experiment in \cite{wu2018group}. However, limited by the computational resources, the experiment is conducted on CIFAR-10, CIFAR-100 \cite{krizhevsky2009learning}, and SVHN \cite{svhn2011netzer} dataset for the image classification task (see Appendix \ref{app:experiment_details} for the details of the experiment settings and the hyperparameter). We set the number of groups (G) with 32 in all GN-based calculations. In the experiment, we exclude BN as the best result from the comparison even though it achieves the best performance in all datasets. It is because our proposed method aims to improve GN and address the issues in GN.

\subsection{Image Classification in CIFAR-10}

\begin{table}[ht]
    \centering
    \begin{tabular}{c c c c c c}
    \toprule
    \textbf{Batch Size} & BN & GN & A & B & C\\
    \midrule
        8 & 95.17 & \textbf{94.01} & - & 94 & - \\
        16 & 95.38 & 93.74 & - & \textbf{93.82} & - \\
        32 & 95.31 & \textbf{93.52} & 86.1 & 92.5 & 94.62 \\
        64 & 94.56 & 92.66 & 93.48 & \textbf{93.57} & 93.39 \\
        128 & 94.19 & 91.88 & 92.27 & \textbf{92.39} & 92.4 \\
    \midrule
        Mean & 94.92 & 93.16 & 90.62 & \textbf{93.26} & 93.47 \\
        Variance & 0.27 & 0.77 & 15.67 & \textbf{0.57} & 1.24 \\
    \bottomrule
    \end{tabular}
    \caption{Validation accuracy (\%) of different normalization layers in CIFAR-10 dataset. A, B, C denotes our proposed method which are GNPlusBNFirst, GNPlusGNFirst, and GNPlusParallel correspondingly. Variant A and C are excluded from the best result because gradient explosion occurs for this variants.}
    \label{table:cifar_10_result}
\end{table}

In Table \ref{table:cifar_10_result}, we report different normalization layers' performance trained with varying batch sizes. Our proposed solution outperforms GN denoted in higher mean and lower variance for the validation accuracy. The improvement may come from better optimization of the network that uses the proposed normalization layer. Since the layer can choose the best option (between using BN or GN results), it can maximize the optimization. As a result, the validation accuracy also becomes better. However, our proposed solution variant such as GNPlusBNFirst and GNPlusParallel suffers from the gradient explosion problem. This shows an interesting observation that applying GN calculation to BN-normalized output can introduce instability in the training process. In addition, even though GNPlusParallel is the variant that can choose whether to use BN or GN output (both uses $x$ as the input) as the output of the normalization layer, this layer also suffers from gradient explosion. This result indicates that applying BN calculation to GN-normalized output like in the GNPlusGNFirst layer can stabilize the training process better. This observation shows that BN normalization indeed can stabilize and help the training process of neural networks, which is in accordance with the study in \cite{santurkar2018how}.

\subsection{Image Classification in CIFAR-100 and SVHN}

\begin{table}[ht]
    \centering
    \begin{tabular}{c c c c}
    \toprule
    \textbf{Batch Size} & BN & GN & GNPlusGNFirst\\
    \midrule
        8 & 78.7 & \textbf{75.33} & 74.81 \\
        16 & 79.87 & 74.83 & \textbf{77.17} \\
        32 & 78.9 & 71.32 & \textbf{76.86} \\
        64 & 78.04 & 71.79 & \textbf{75.21} \\
        128 & 77.85 & 63.88 & \textbf{72.88} \\
    \midrule
        Mean & 78.67 & 71.43 & \textbf{75.39} \\
        Variance & 0.64 & 20.98 & \textbf{3.00} \\
    \bottomrule
    \end{tabular}
    \caption{Validation accuracy (\%) of different normalization layers in CIFAR-100 dataset.}
    \label{table:cifar_100_result}
\end{table}

\begin{table}[ht]
    \centering
    \begin{tabular}{c c c c}
    \toprule
    \textbf{Batch Size} & BN & GN & GNPlusGNFirst\\
    \midrule
        8 & 96.73 & \textbf{96.54} & - \\
        16 & 96.63 & \textbf{96.49} & 95.95 \\
        32 & 96.52 & \textbf{96.34} & 96.22 \\
        64 & 96.25 & \textbf{96.17} & 95.33 \\
        128 & 96.52 & 91.80 & \textbf{94.73} \\
    \midrule
        Mean & 96.53 & 95.47 & \textbf{95.56} \\
        Variance & 0.03 & 4.22 & \textbf{0.44} \\
    \bottomrule
    \end{tabular}
    \caption{Validation accuracy (\%) of different normalization layers in SVHN dataset.}
    \label{table:svhn_result}
\end{table}

In this experiment, we only show GNPlusGNFirst as the representative of our proposed normalization layer because other variants are unstable. The validation accuracy when the network is trained with different batch size settings on CIFAR-100 and SVHN can be seen in Table \ref{table:cifar_100_result} and Table \ref{table:svhn_result} respectively. In the CIFAR-100 result, we observe that GNPlusGNFirst outperforms GN in both variance and mean by a high margin. The low variance that the proposed method achieved implies that our solution has better stability to the choice of training batch size compared to GN. 

As for the experiment on the SVHN dataset, our method suffers from a gradient explosion problem when trained using a small batch size setting. The explosion likely occurs due to the instability of the method. Recall that our method employs $\lambda$ to control whether to use GN or BN output. If $\lambda$ changes rapidly (because both feature statistics are unreliable), then it will introduce some instability in the training process which causes the gradient explosion problem. We can improve our method by employing some regularization where we limit the $\lambda$ rate of change. Nevertheless, if we exclude the result where gradient explosion occurs then our method again outperforms GN both in the stability with respect to batch size variable (variance) and the overall performance (mean) that our method achieves.

In conclusion, after analyzing the result of all experiments across different datasets, we find that BN performance is insensitive to the choice of batch size (low variance across experiments) in contrary to the study in \cite{wu2018group}. Meanwhile, we find that GN performance is sensitive to the choice of batch size. This is shown by a tendency where GN performance constantly decreases when the batch size is increased for all of the experiments. However, the result of our proposed method shows better performance and also lesser sensitivity to the choice of batch size.

\subsection{Impact of Proposed Normalization Layer on The Neural Network Training}

The impact (measured by loss landscape and gradient predictiveness) of the proposed normalization layer (GNPlusGNFirst variant) in helping to ease the network training are shown in Figure \ref{fig:appendix_gn_plus_gn_first_loss_landscape_grad_pred}. We observe that both loss landscape and gradient predictiveness of the proposed layer have a smaller range of values and lower values compared to the original GN. It demonstrates that \textbf{the proposed layer achieves better stability in training}. These results show that our method built on top of GN by modifying the underlying mechanism of the GN layer successfully enhances the ability of the layer to help the training process of the neural network. As a consequence, the GNPlusGNFirst layer also becomes more robust to the choice of hyperparameter. Therefore, it can prevent gradient vanishing or explosion problems better.

The \textbf{robustness} of the proposed layer to the noise induced by the regularization effect is shown by the loss landscape and gradient predictiveness which can be seen in Figure \ref{fig:appendix_regularization_gn_plus_gn_first_loss_landscape_grad_pred}. Both metrics have a stable result (does not change) even though regularization is applied. The result is in contrast to the case of using GN as the normalization layer, where the loss landscape and gradient predictiveness becomes wider when regularization is introduced in the training. 

In addition, the result that demonstrates the robustness of the proposed layer to the distortion of the output distribution can be seen in Figure \ref{fig:appendix_gn_plus_gn_first_under_distortion}. We observe that the network with our proposed layer can be safely trained despite the distortion in the output distribution of the layer caused by the Gaussian noise. Interestingly, the network with the proposed layer also achieves the best validation performance compared to BN and GN.

To summarize, we conclude that our proposed normalization layer builds on top of GN can solve all of the issues that GN has. The main reason behind the improvements is because \textbf{we incorporate BN into the GN layer that smooths the underlying optimization problem}. Moreover, after conducting various experiments, we can also conclude that the proposed layer is superior to the vanilla GN layer if we exclude one failure case where our method suffers from the gradient explosion.

\section{Conclusion and Future Works}
In this paper, we build an analysis framework to analyze the GN's working principles and explore the root cause (e.g., issues) of its inferior performance against BN. To summarize, we conclude that, first, GN is sensitive to the distortion of the output distribution and the perturbation induced by the effect of regularization. Second, GN can only help the training of the neural network slightly in the middle period of training. Last, the layer makes a negative influence on the initial period of the neural network training. To solve these issues, we propose three new normalization layer variants build on top of GN. The main idea is to incorporate BN calculation adaptively to the GN layer in a sequential or parallel manner. We find that our new normalization layer, especially GNPlusGNFirst achieves better performance than GN in the experiment across different datasets. In addition, our method also solves all of the issues that GN has.

Despite the better performance that our proposed layer achieves than the original GN layer, we realize that our method (GNPlusGNFirst) still brings a risk of gradient explosion in certain settings (see the experiments on SVHN). This issue can be possibly solved by introducing extra regularization terms to the parameter $\lambda$. In the future, we would evaluate the proposed layer in larger-size datasets (e.g., ImageNet) and apply it to other vision tasks, such as object detection and video classification. In addition, we will also learn other theoretical studies to understand the exact effect of the GN layer on network optimization, hoping to improve GN further.

{\small
\bibliographystyle{ieee_fullname}
\bibliography{egbib}

\begin{thebibliography}{10}\itemsep=-1pt

\bibitem{ba2016layer}
Jimmy~Lei Ba, Jamie~Ryan Kiros, and Geoffrey~E Hinton.
\newblock Layer normalization.
\newblock {\em arXiv preprint arXiv:1607.06450}, 2016.

\bibitem{goodfellow2014generative}
Ian~J Goodfellow, Jean Pouget-Abadie, Mehdi Mirza, Bing Xu, David Warde-Farley,
  Sherjil Ozair, Aaron Courville, and Yoshua Bengio.
\newblock Generative adversarial networks.
\newblock {\em arXiv preprint arXiv:1406.2661}, 2014.

\bibitem{he2016deep}
Kaiming He, Xiangyu Zhang, Shaoqing Ren, and Jian Sun.
\newblock Deep residual learning for image recognition.
\newblock In {\em Proceedings of the IEEE conference on computer vision and
  pattern recognition}, pages 770--778, 2016.

\bibitem{hochreiter1997long}
Sepp Hochreiter and J{\"u}rgen Schmidhuber.
\newblock Long short-term memory.
\newblock {\em Neural computation}, 9(8):1735--1780, 1997.

\bibitem{huang2017densely}
Gao Huang, Zhuang Liu, Laurens Van Der~Maaten, and Kilian~Q Weinberger.
\newblock Densely connected convolutional networks.
\newblock In {\em Proceedings of the IEEE conference on computer vision and
  pattern recognition}, pages 4700--4708, 2017.

\bibitem{huang2017arbitrary}
Xun Huang and Serge Belongie.
\newblock Arbitrary style transfer in real-time with adaptive instance
  normalization.
\newblock In {\em Proceedings of the IEEE International Conference on Computer
  Vision}, pages 1501--1510, 2017.

\bibitem{ioffe2017renormalization}
Sergey Ioffe.
\newblock Batch renormalization: Towards reducing minibatch dependence in
  batch-normalized models.
\newblock In I. Guyon, U.~V. Luxburg, S. Bengio, H. Wallach, R. Fergus, S.
  Vishwanathan, and R. Garnett, editors, {\em Advances in Neural Information
  Processing Systems}, volume~30. Curran Associates, Inc., 2017.

\bibitem{ioffe2015batch}
Sergey Ioffe and Christian Szegedy.
\newblock Batch normalization: Accelerating deep network training by reducing
  internal covariate shift.
\newblock In {\em International conference on machine learning}, pages
  448--456. PMLR, 2015.

\bibitem{kingma2015adam}
Diederik~P. Kingma and Jimmy Ba.
\newblock Adam: {A} method for stochastic optimization.
\newblock In Yoshua Bengio and Yann LeCun, editors, {\em 3rd International
  Conference on Learning Representations, {ICLR} 2015, San Diego, CA, USA, May
  7-9, 2015, Conference Track Proceedings}, 2015.

\bibitem{krizhevsky2009learning}
Alex Krizhevsky, Geoffrey Hinton, et~al.
\newblock Learning multiple layers of features from tiny images.
\newblock 2009.

\bibitem{li2018visualizing}
Hao Li, Zheng Xu, Gavin Taylor, Christoph Studer, and Tom Goldstein.
\newblock Visualizing the loss landscape of neural nets.
\newblock In S. Bengio, H. Wallach, H. Larochelle, K. Grauman, N. Cesa-Bianchi,
  and R. Garnett, editors, {\em Advances in Neural Information Processing
  Systems}, volume~31. Curran Associates, Inc., 2018.

\bibitem{lian2019revisit}
Xiangru Lian and Ji Liu.
\newblock Revisit batch normalization: New understanding and refinement via
  composition optimization.
\newblock In Kamalika Chaudhuri and Masashi Sugiyama, editors, {\em Proceedings
  of the Twenty-Second International Conference on Artificial Intelligence and
  Statistics}, volume~89 of {\em Proceedings of Machine Learning Research},
  pages 3254--3263. PMLR, 16--18 Apr 2019.

\bibitem{luo2018towards}
Ping Luo, Xinjiang Wang, Wenqi Shao, and Zhanglin Peng.
\newblock Towards understanding regularization in batch normalization.
\newblock In {\em International Conference on Learning Representations}, 2019.

\bibitem{svhn2011netzer}
Yuval Netzer, Tao Wang, Adam Coates, Alessandro Bissacco, Bo Wu, and Andrew~Y.
  Ng.
\newblock Reading digits in natural images with unsupervised feature learning.
\newblock In {\em NIPS Workshop on Deep Learning and Unsupervised Feature
  Learning 2011}, 2011.

\bibitem{santurkar2018how}
Shibani Santurkar, Dimitris Tsipras, Andrew Ilyas, and Aleksander Madry.
\newblock How does batch normalization help optimization?
\newblock In S. Bengio, H. Wallach, H. Larochelle, K. Grauman, N. Cesa-Bianchi,
  and R. Garnett, editors, {\em Advances in Neural Information Processing
  Systems}, volume~31. Curran Associates, Inc., 2018.

\bibitem{simonyan2015vgg}
Karen Simonyan and Andrew Zisserman.
\newblock Very deep convolutional networks for large-scale image recognition.
\newblock In Yoshua Bengio and Yann LeCun, editors, {\em 3rd International
  Conference on Learning Representations, {ICLR} 2015, San Diego, CA, USA, May
  7-9, 2015, Conference Track Proceedings}, 2015.

\bibitem{summers2020four}
Cecilia Summers and Michael~J. Dinneen.
\newblock Four things everyone should know to improve batch normalization.
\newblock In {\em International Conference on Learning Representations}, 2020.

\bibitem{szegedy2016rethinking}
Christian Szegedy, Vincent Vanhoucke, Sergey Ioffe, Jon Shlens, and Zbigniew
  Wojna.
\newblock Rethinking the inception architecture for computer vision.
\newblock In {\em Proceedings of the IEEE conference on computer vision and
  pattern recognition}, pages 2818--2826, 2016.

\bibitem{ulyanov2016instance}
Dmitry Ulyanov, Andrea Vedaldi, and Victor Lempitsky.
\newblock Instance normalization: The missing ingredient for fast stylization.
\newblock {\em arXiv preprint arXiv:1607.08022}, 2016.

\bibitem{wu2018group}
Yuxin Wu and Kaiming He.
\newblock Group normalization.
\newblock In {\em Proceedings of the European conference on computer vision
  (ECCV)}, pages 3--19, 2018.

\bibitem{jingjing2019understanding}
Jingjing Xu, Xu Sun, Zhiyuan Zhang, Guangxiang Zhao, and Junyang Lin.
\newblock Understanding and improving layer normalization.
\newblock In H. Wallach, H. Larochelle, A. Beygelzimer, F. d\textquotesingle
  Alch\'{e}-Buc, E. Fox, and R. Garnett, editors, {\em Advances in Neural
  Information Processing Systems}, volume~32. Curran Associates, Inc., 2019.

\bibitem{zhou2020batch}
Xiao-Yun Zhou, Jiacheng Sun, Nanyang Ye, Xu Lan, Qijun Luo, Bo-Lin Lai, Pedro
  Esperanca, Guang-Zhong Yang, and Zhenguo Li.
\newblock Batch group normalization.
\newblock {\em arXiv preprint arXiv:2012.02782}, 2020.

\end{thebibliography}
}

\newpage
\appendix
\section{Omitted Experiments}
All of the experiments that are omitted from the report can be seen in the following subsections.

\subsection{Group Normalization (GN) Applicability to Different Neural Network Families}
\label{app:exp_gn_aplicability}

\begin{figure}[ht]
\begin{center}
\includegraphics[width=\columnwidth]{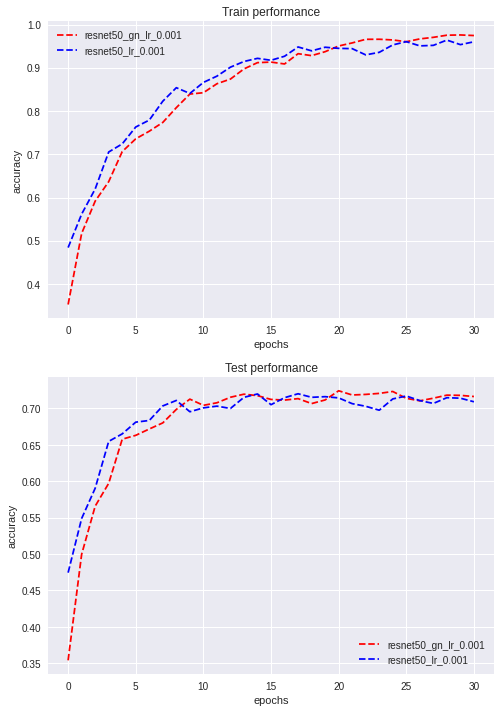}
\end{center}
   \caption{Training and validation performance of ResNet50 using GN as the normalization layer compared to the vanilla network.}
\label{fig:resnet50_gn}
\end{figure}

\begin{figure}[ht]
\begin{center}
\includegraphics[width=\columnwidth]{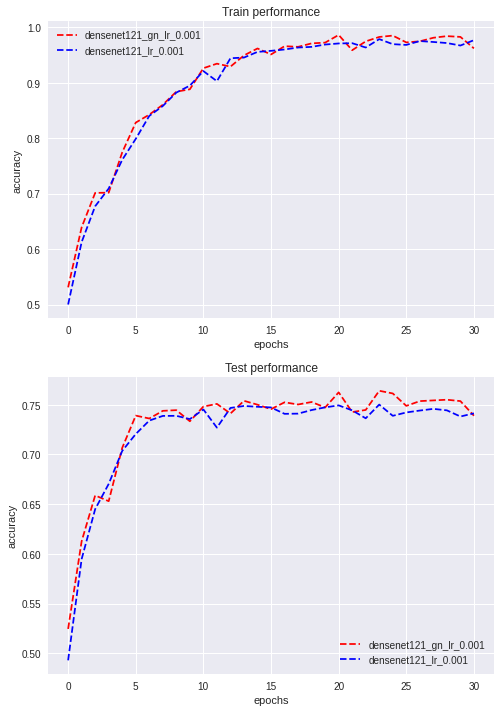}
\end{center}
   \caption{Training and validation performance of DenseNet121 using GN as the normalization layer compared to the vanilla network.}
\label{fig:densenet121_gn}
\end{figure}

\begin{figure}[ht]
\begin{center}
\includegraphics[width=\columnwidth]{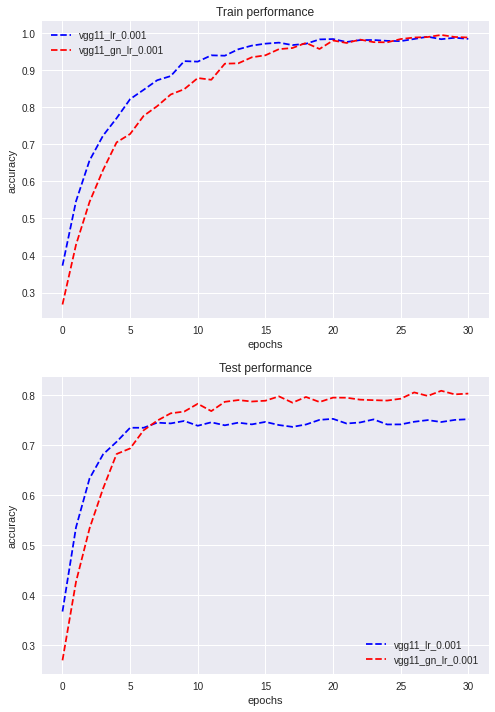}
\end{center}
   \caption{Training and validation performance of VGG-11 using GN as the normalization layer compared to the vanilla network.}
\label{fig:vgg11_gn}
\end{figure}

In this experiment, we use GN in different neural network families that have different learning behavior such as ResNet \cite{he2016deep} which employs residual learning, DenseNet \cite{huang2017densely} which employs dense connection to help training, and VGG \cite{simonyan2015vgg} which can be trained without any normalization layer. As shown in Figure \ref{fig:resnet50_gn}, Figure \ref{fig:densenet121_gn}, and Figure \ref{fig:vgg11_gn}, GN helps in easing the training of different neural networks family and also improves the validation performance of them.

\subsection{Sensitivity of Group Normalization (GN) to The Distortion in The Output Distribution}
\label{app:exp_gn_sensitivity_to_distortion}
\begin{figure}[ht]
\begin{center}
\includegraphics[width=\columnwidth]{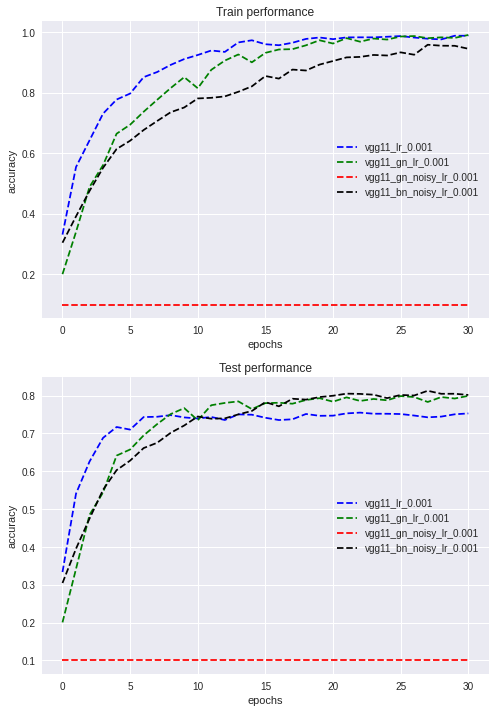}
\end{center}
   \caption{Training and validation performance of VGG-11 using different normalization layers. The result of adding Gaussian noise to the output of the normalization layer can be seen in red lines and black lines for the network with GN and BN correspondingly.}
\label{fig:appendix_gn_with_distortion}
\end{figure}

In this experiment, we train a VGG-11 network for 50 epochs using the CIFAR-10 dataset. The learning rate used for this experiment is $10^{-3}$ where Adam is employed as the optimizer. For the 'noisy' result, we inject a Gaussian noise with $\mu = 10^{-3}$ and $\sigma = 1.001$ for each output of the normalization layer. As shown in Figure \ref{fig:appendix_gn_with_distortion}, when the network uses GN for the normalization layer and some noise is added to distort the output distribution, the network suffers from a gradient vanishing problem that makes further training is futile. Meanwhile, adding some noise to the output of BN does not make the network suffers from this problem. This means that GN is sensitive to the distortion of output distribution compared to BN.

\subsection{Loss Landscape and Gradient Predictiveness of Group Normalization (GN)}
\label{app:omitted_exp_gn_landscape_and_grad_pred}

\begin{figure*}[ht]
\begin{center}
\includegraphics[width=\columnwidth]{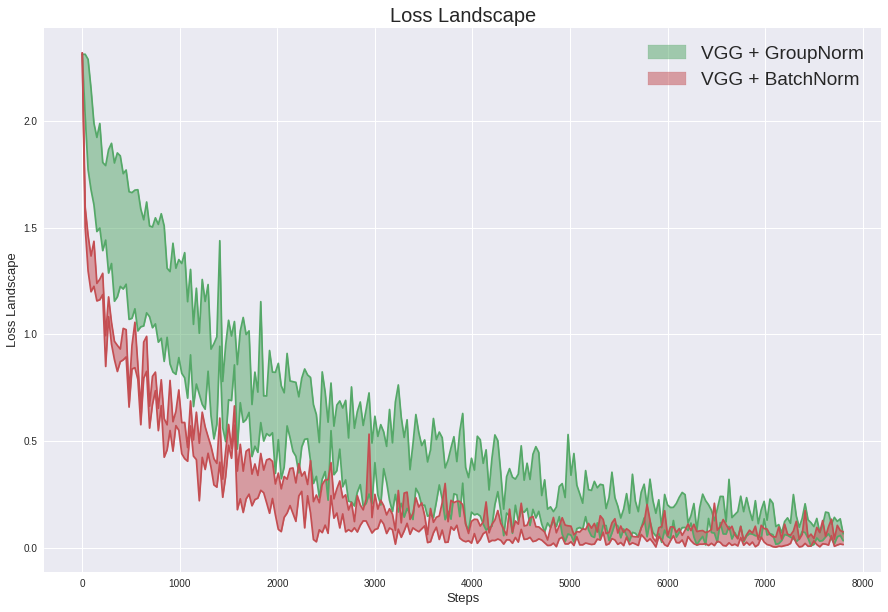}
\includegraphics[width=\columnwidth]{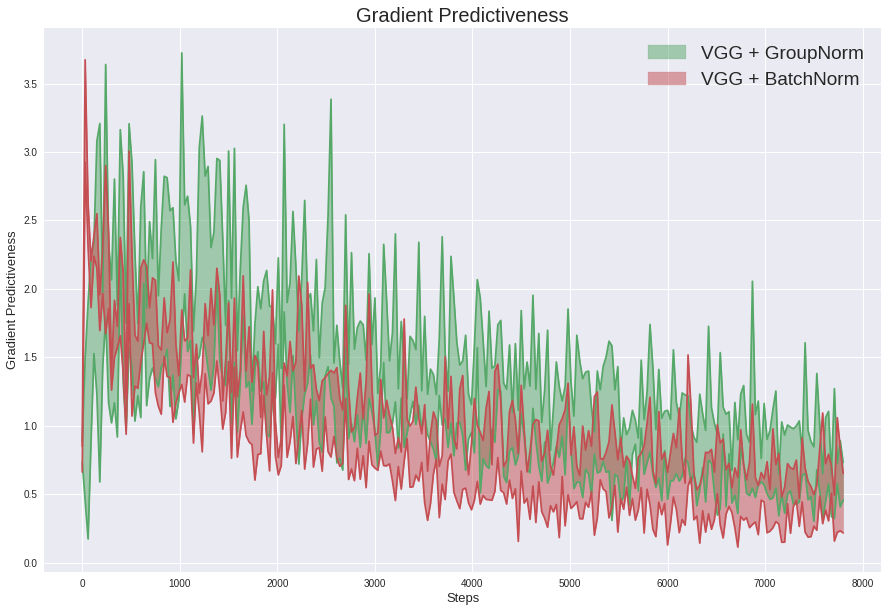}
\end{center}
   \caption{Loss landscape (left) and gradient predictiveness (right) of the network that uses GN (green) as the normalization layer compared to BN (red).}
\label{fig:appendix_gn_vs_bn_loss_landscape_grad_pred}
\end{figure*}

\begin{figure*}[ht]
\begin{center}
\includegraphics[width=\columnwidth]{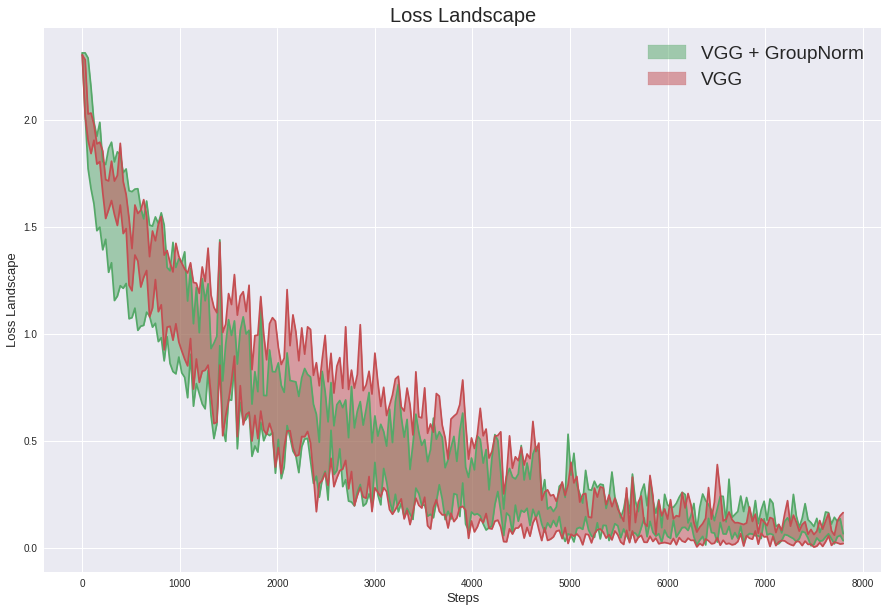}
\includegraphics[width=\columnwidth]{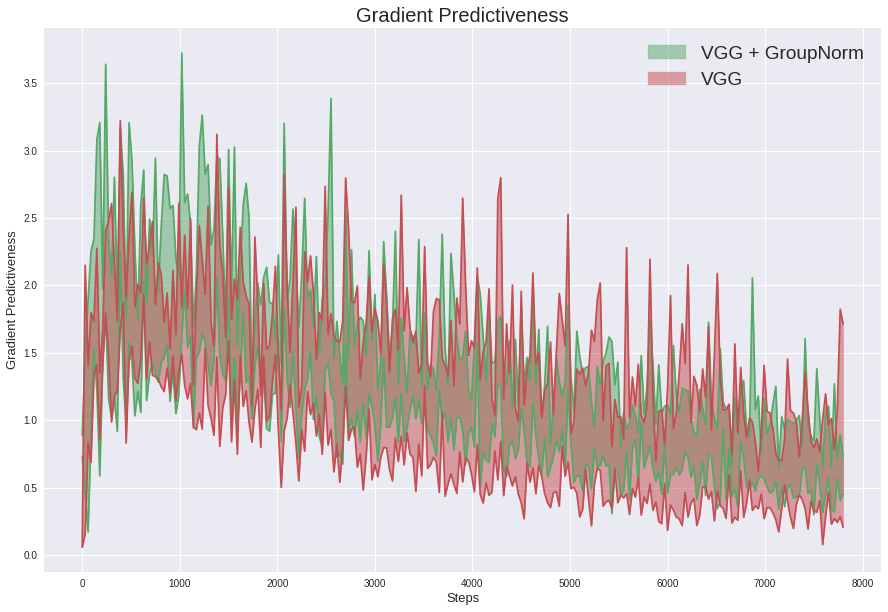}
\end{center}
   \caption{Loss landscape (left) and gradient predictiveness (right) of the network that uses GN (green) as the normalization layer compared to the network without any normalization layer (red).}
\label{fig:appendix_loss_landscape_grad_pred_gn_vs_nonorm}
\end{figure*}

In this experiment, we train a VGG-11 network for 20 epochs with five different learning rates chosen uniformly from $10^{-4}$ until $5 \times 10^{-4}$ with an increment of $10^{-4}$. In Figure \ref{fig:appendix_gn_vs_bn_loss_landscape_grad_pred} the loss landscape and gradient predictiveness of the network which uses GN as the normalization layer compared to BN can be seen. We can observe that both loss landscape and gradient predictiveness of the network that uses BN as the normalization layer are better. The loss landscape of the network with BN has a lower value and lower range of values which denotes better stability of the optimization. Meanwhile, the gradient predictiveness of BN also has a lower value and lower range of values that again indicate better stability of optimization. Specifically, it shows better stability and predictiveness of the gradients.

Figure \ref{fig:appendix_loss_landscape_grad_pred_gn_vs_nonorm} shows the loss landscape and gradient predictiveness of GN compared to no normalization layer. The result shows that in the initial training period, the network without the normalization layer has better stability in the network optimization (network training) compared to GN. This means that GN can disrupt the training process of the neural network, especially in the early period. In the middle period of training (e.g. after 1000 training steps and before starting to converge), GN helps to ease the training of the neural network better denoted by a lower range of values in the loss landscape and also a lower range of values in the gradient predictiveness. Based on these results, we can conclude that the effect of GN on neural network training is positive on the middle period of training and negative on the initial period of training.

\subsection{Loss Landscape and Gradient Predictiveness of Group Normalization (GN) Under Regularization Effect}

\begin{figure*}[ht]
\begin{center}
\includegraphics[width=\columnwidth]{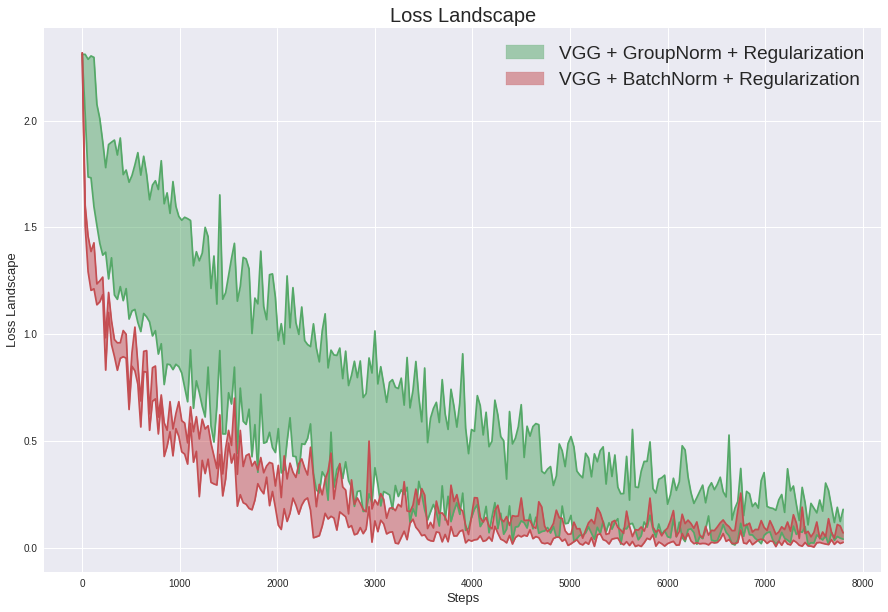}
\includegraphics[width=\columnwidth]{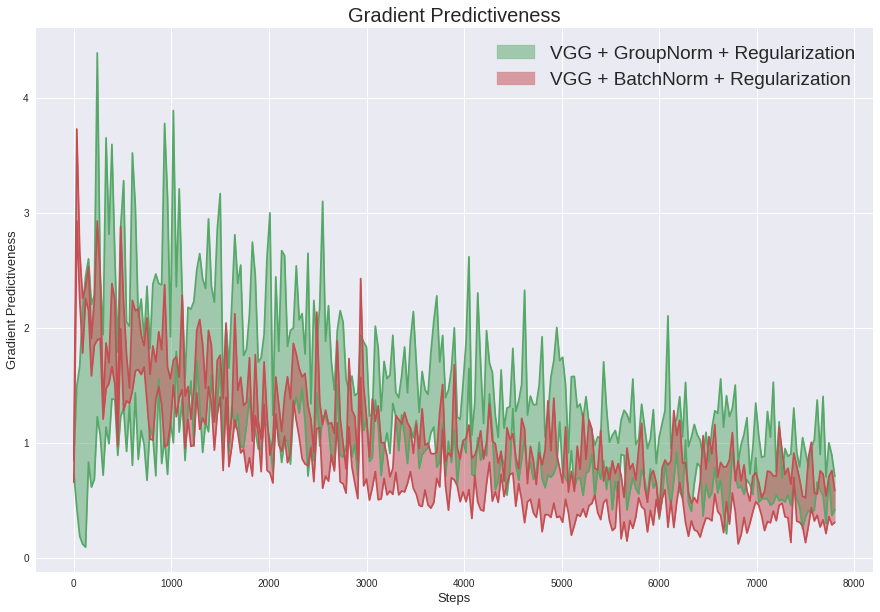}
\end{center}
   \caption{Loss landscape (left) and gradient predictiveness (right) of the network that uses GN (green) and BN (red) trained with L2-regularization.}
\label{fig:appendix_regularization_loss_landscape_grad_pred_gn_vs_bn}
\end{figure*}

In this experiment, we train a VGG-11 network for 20 epochs using the same hyperparameters that have been explained in Appendix \ref{app:analysis_experiments_details} and L2-regularization (weight decay) with a magnitude of $5 \times 10^{-5}$. The loss landscape and gradient predictiveness of the network with GN as the normalization layer compared to BN trained under L2-regularization can be seen in Figure \ref{fig:appendix_regularization_loss_landscape_grad_pred_gn_vs_bn}. As shown in the figure, for both the loss landscape and gradient predictiveness, the range of values of the network with GN is increased (compared with Figure \ref{fig:appendix_gn_vs_bn_loss_landscape_grad_pred}) when the regularization is applied to the network training. Meanwhile, regularization does not affect the training process of the neural network with the BN layer shown by its loss landscape and gradient predictiveness that does not change significantly even though regularization is applied in the training.

\subsection{Evaluation of Other Ideas}
\label{app:omitted_exp_evaluation_of_other_ideas}

In this experiment, we evaluate other ideas of combining both GN and BN in \cite{summers2020four} and \cite{zhou2020batch}. In addition, we also evaluate the idea of combining GN and BN by stacking them sequentially. In \cite{summers2020four}, the author proposes General Batch and Group Normalization (GBGN) which uses the grouping mechanism for both channels and examples. Meanwhile, in \cite{zhou2020batch}, Batch Group Normalization (BGN) is proposed to group the channels, height, and width dimension. However, in our exploratory study, we find that both these methods and adding BN layer directly after GN layer suffer from gradient vanishing problem. For this exploratory study, we train VGG-11 using the CIFAR-10 dataset with Adam optimizer. The test cases where BN can be safely trained but GN suffers from gradient vanishing problem are in the settings of A: $lr=0.01, batch\_size=128$, and B: $lr=0.001, batch\_size=4$. We evaluate GBGN by using 2 as the group number for the examples. This means for a batch size of 128, the number of examples per group is 64. The evaluation shows that GBGN can be used successfully in setting A but fails in setting B. Meanwhile, we employ 32 as the number of groups in BGN. The result in BGN also has similar cases where it fails in setting B.

\subsection{Loss Landscape and Gradient Predictiveness of GNPlusGNFirst Layer}

\begin{figure*}[ht]
\begin{center}
\includegraphics[width=\columnwidth]{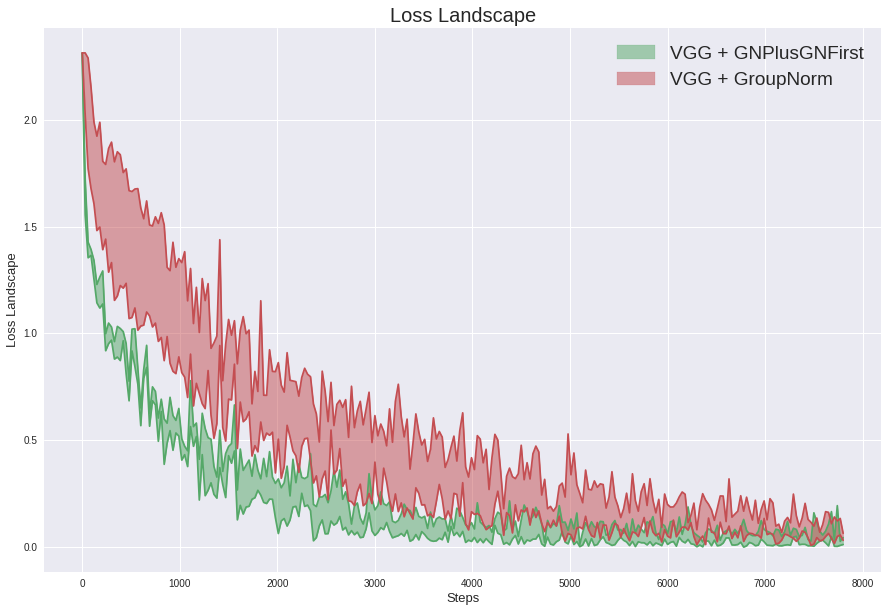}
\includegraphics[width=\columnwidth]{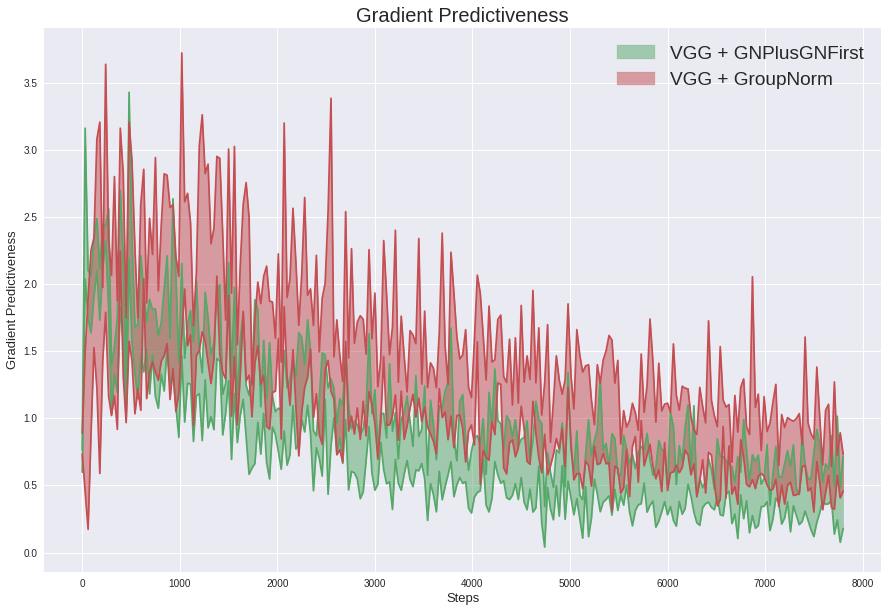}
\end{center}
   \caption{Loss landscape (left) and gradient predictiveness (right) of the network that uses GNPlusGNFirst (green) as the normalization layer compared to GN (red).}
\label{fig:appendix_gn_plus_gn_first_loss_landscape_grad_pred}
\end{figure*}

\begin{figure*}[ht]
\begin{center}
\includegraphics[width=\columnwidth]{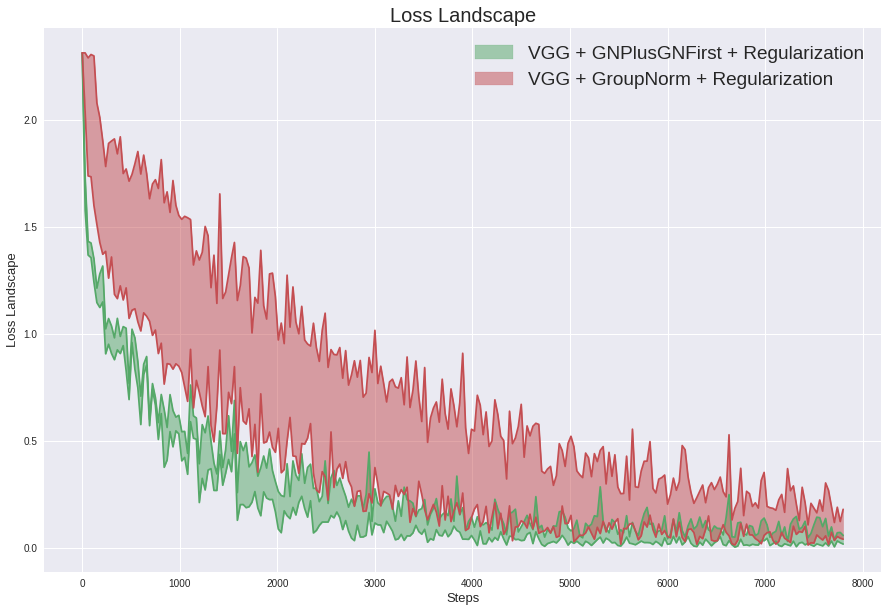}
\includegraphics[width=\columnwidth]{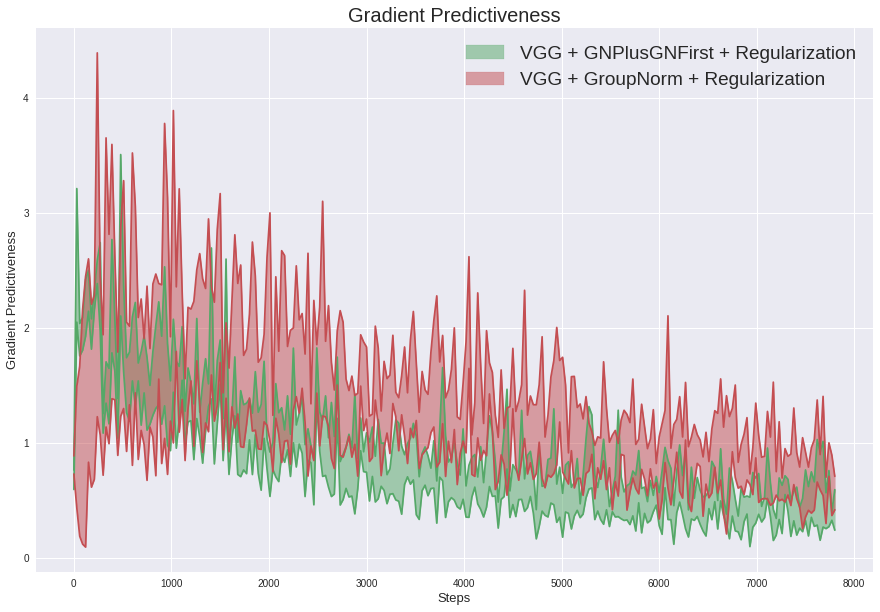}
\end{center}
   \caption{Loss landscape (left) and gradient predictiveness (right) of the network that uses GNPlusGNFirst (green) as the normalization layer compared to GN (red) when regularization (L2-regularization with a magnitude of $5 \times 10^{-5}$) is applied.}
\label{fig:appendix_regularization_gn_plus_gn_first_loss_landscape_grad_pred}
\end{figure*}

The loss landscape and gradient predictiveness of the proposed normalization layer (GNPlusGNFirst) compared to GN can be seen in Figure \ref{fig:appendix_gn_plus_gn_first_loss_landscape_grad_pred}. We can observe that our proposed normalization layer has a lower range of values and lower value for the loss landscape. This denotes that our proposed normalization layer helps the optimization of the network better compared to GN. In addition, the gradient predictiveness also shows similar observation where the proposed layer has lower values and a lower range of values which further supports that the proposed normalization layer has a better impact on the training process (optimization) of the network.

Moreover, the loss landscape and gradient predictiveness of the proposed normalization layer compared to GN under the effect of regularization can be seen in Figure \ref{fig:appendix_regularization_gn_plus_gn_first_loss_landscape_grad_pred}. If we compare with the result of Figure \ref{fig:appendix_gn_plus_gn_first_loss_landscape_grad_pred}, then both the loss landscape and gradient predictiveness of the proposed normalization layer does not change significantly compared to GN. This shows that our proposed normalization layer has better robustness and is not sensitive to the noise introduced by regularization compared to GN.

\subsection{Sensitivity of GNPlusGNFirst to The Distortion in The Output Distribution}
\label{app:omitted_exp_gn_plus_gn_first}

\begin{figure}[ht]
\begin{center}
\includegraphics[width=\columnwidth]{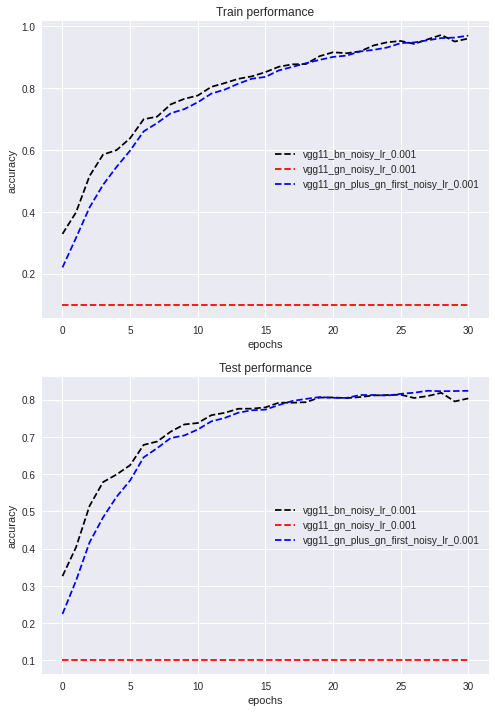}
\end{center}
   \caption{Training and validation performance of VGG-11 using different normalization layers under the distortion of the output distribution. Black, red, and blue lines denote the result of using BN, GN, and GNPlusGNFirst respectively.}
\label{fig:appendix_gn_plus_gn_first_under_distortion}
\end{figure}

In this experiment, we train the VGG-11 network using the CIFAR-10 dataset with an additional injection of noise for each normalized output. The noise is a Gaussian noise with $\mu = 10^{-3}$ and $\sigma = 1.001$. The learning rate used in this experiment is $10^{-3}$ and Adam is employed as the optimizer. As can be seen in Figure \ref{fig:appendix_gn_plus_gn_first_under_distortion}, the network that uses the proposed layer (GNPlusGNFirst) can be safely trained and even achieves the best validation performance compared to BN and GN.

\section{Omitted Details}

\subsection{Experiment Settings}
\label{app:experiment_details}

The hyperparameter used for training with different datasets (CIFAR-10, CIFAR-100, SVHN) can be described by the following points.
\begin{itemize}
    \item learning rate: for all of the dataset, the learning rate use the formulation of $10^{-1} \times (batch\_size / 128)$.
    \item epochs: for all of the datasets, the network is trained for 164 epochs.
    \item learning rate decay schedule: for all of the dataset, the learning rate is decreased by a factor of 10 after 81 and 122 epochs.
    \item weight decay: the weight decay used for the CIFAR-10 dataset is $10^{-4}$, while the weight decay used for CIFAR-100 and SVHN dataset are $5 \times 10^{-4}$.
\end{itemize}

\subsection{Analysis Experiments Details}
\label{app:analysis_experiments_details}

In all of the analysis experiments, we use CIFAR-10 as the training data. To simulate the movement to a different point in the training process, we train the network using 5 different learning rates. These 5 learning rates are picked from $10^{-4}$ until $5 \times 10^{-4}$ with the increment of $10^{-4}$. We employ Adam \cite{kingma2015adam} as the optimizer and use Cross-Entropy as the loss function. The training batch size used throughout the experiment is 128. In all of the experiments, 32 is chosen as the group hyperparameter for GN.

\end{document}